\documentclass[journal]{IEEEtran}
\usepackage{amsmath, amsfonts, amssymb, amsthm}
\usepackage{xcolor}
\usepackage{siunitx}
\usepackage{hyperref}
\hypersetup{colorlinks,allcolors=blue}
\usepackage{algorithm}        
\usepackage{algpseudocode}
\usepackage{gensymb}   
\usepackage{array}
\usepackage{multirow}
\usepackage{booktabs}
\usepackage{tabularx}
\renewcommand{\arraystretch}{1.15}
\newcolumntype{C}[1]{>{\centering\arraybackslash}p{#1}}
\newcolumntype{L}[1]{p{#1}}
\newcolumntype{Y}{>{\centering\arraybackslash}X}

\usepackage{multidef}
\usepackage{mathtools}

\usepackage{xcolor}

\usepackage{amsthm}
\usepackage{comment}
\usepackage{bm}
\usepackage{multidef}
\usepackage{caption}
\usepackage{subcaption}
\usepackage{comment}
\usepackage{adjustbox}
\usepackage{amsfonts,amssymb}
\usepackage{multirow}
\usepackage{nicematrix}
\usepackage{booktabs}
\usepackage{lscape}
\usepackage{optidef}
\usepackage{algorithm}
\usepackage{algpseudocode}
\usepackage{graphicx}
\multidef[prefix=v]{\ensuremath{\mathbf{#1}}}{a-z}

\multidef[prefix=set]{\ensuremath{\mathbb{#1}}}{A-Z}
\multidef[prefix=m]{\ensuremath{\mathbf{#1}}}{A-Z}

\newcommand{\calX}{\mathcal{X}}

\newcommand{\calS}{\mathcal{S}}

\newcommand{\calY}{\mathcal{Y}}
\newcommand{\calW}{\mathcal{W}}
\newcommand{\calB}{\mathcal{B}}
\newcommand{\calL}{\mathcal{L}}

\newcommand{\Bt}{\calW_t^{k+1}-\frac{\Lambda_t^k}{\rho}}
\newcommand{\Bl}{\calW_l^{k+1}-\frac{\Lambda_l^k}{\rho}}
\newcommand{\Bone}{\calW^{k+1}-\frac{\Lambda^k}{\rho}}
\newcommand{\Bb}{\calY-\calX^{k+1}-\frac{\Lambda_f^k}{\rho}}

\newtheorem{assumption}{Assumption}

\theoremstyle{plain}
\newtheorem{theorem}{Theorem}[section]
\newtheorem*{theorem*}{Theorem}
\newtheorem{lemma}[theorem]{Lemma}

\theoremstyle{definition}
\newtheorem{definition}{Definition}

\theoremstyle{remark}

\numberwithin{theorem}{section}

\begin{document}

\markboth{}{Author {et al.}}

\title{Robust Spatiotemporally Contiguous Anomaly Detection Using Tensor Decomposition}
\author{%
Rachita Mondal$^{1}$, 
Mert Indibi$^{2}$, 
Tapabrata Maiti$^{1}$, 
and Selin Aviyente$^{2}$, \textit{Senior Member, IEEE}%
\thanks{$^{1}$Department of Statistics and Probability, Michigan State University, East Lansing, MI, 48824}%
\thanks{$^{2}$Department of Electrical and Computer Engineering, Michigan State University, East Lansing, MI, 48824}%
\thanks{This work was supported in part by the National Science Foundation under Grants CCF-2211645 and CCF-2312546.}
}



\maketitle
\thispagestyle{empty}
\begin{abstract}
 Anomaly detection in spatiotemporal data is a challenging problem encountered in a variety of applications, including video surveillance, medical imaging data, and urban traffic monitoring. Existing anomaly detection methods focus mainly on point anomalies and cannot deal with temporal and spatial dependencies that arise in spatio-temporal data. Tensor-based anomaly detection methods have been proposed to address this problem. Although existing methods can capture dependencies across different modes, they are primarily supervised and do not account for the specific structure of anomalies. Moreover, these methods focus mainly on extracting  anomalous features without providing any statistical confidence. In this paper, we introduce an unsupervised tensor-based anomaly detection method that simultaneously considers the sparse and spatiotemporally smooth nature of anomalies. The anomaly detection problem is formulated as a regularized robust low-rank + sparse tensor decomposition where the total variation of the tensor with respect to the underlying spatial and temporal graphs quantifies the spatiotemporal smoothness of the anomalies. Once the anomalous features are extracted, we introduce a statistical anomaly scoring framework that accounts for local spatio-temporal dependencies. The proposed framework is evaluated on both synthetic and real  data. 
\end{abstract}
\begin{IEEEkeywords}
Anomaly Detection; Spatiotemporal Data; Tensor Decomposition; Graph Total Variation 
\end{IEEEkeywords}

\section{Introduction}\label{introduction}
With recent advances in sensing technology, spatiotemporal (ST) data with rich structural information are acquired for process modeling, monitoring, and diagnosis \cite{bgfg_sep_smooth,yan2014image}. Some examples include dynamic computed tomography (CT) scanning to
illustrate different cross sections of an organ \cite{zhang2016tensor}, geographic information system (GIS) data, video, and traffic flow \cite{tan2016short}. This type of data
is typically collected from multiple spatial locations at regular intervals over time. 
 One of the challenges of analyzing this type of data is detecting anomalies from possibly corrupted observations. 

The problem of anomaly detection has been extensively studied using various methodologies, including statistics \cite{ju2012fuzzy,rousseeuw2003robust}, machine learning \cite{gornitz2013toward,ruff2019deep,villa2021semi}, and spectral algorithms \cite{zhang2016tensor,lin2018anomaly}. However, existing methodology has some limitations when it comes to detecting anomalies in spatiotemporal data. First, obtaining labeled training
data is often too expensive, if not
impossible. Thus, there is a need for unsupervised methods. Second, most current anomaly detection methods focus on point anomalies. However, in spatiotemporal data, group anomalies, which are spatially contiguous groups of locations that show anomalous values consistently for a short duration of time,  are commonly encountered. Some examples of such group anomalies
include traffic accidents that result in abnormally high traffic volume in a given region for a certain duration of time or an abnormal number of tweets  from a spatial region in a small time window \cite{zhang2018detecting,chae2012spatiotemporal}. Therefore, it is important to capture the context in which an anomaly occurs by modeling the spatial and temporal regularity of the data. Finally, most of the existing
techniques lack the capacity to deal with multi-way features, such as those over time and space. A common approach is to decompose the anomaly detection problem, treating the spatial and temporal properties of the outliers independently and then merging them in a post-processing step \cite{wu2008spatio}. As the number of sensors increases, scalability becomes a critical issue and these methods become unreliable.

Tensors, or multi-way arrays, provide a natural representation for spatiotemporal data and low-rank tensor decomposition and completion have been proposed as suitable approaches to anomaly detection as these methods are a natural extension of spectral anomaly detection techniques to multi-way data \cite{zhang2016tensor,li2020tensor,lin2018anomaly,xu2019anomaly,kasai2016network}. These models project the original spatiotemporal data into a low-dimensional latent space, where normal activity is represented with improved spatial and temporal resolution. The learned features, i.e., factor matrices or core tensors, are then used to detect anomalies by monitoring the reconstruction error at each time point \cite{xu2019anomaly,papalexakis2018network} or by applying well-known statistical tests to the extracted multivariate features \cite{zhang2016tensor}. 
Although these low-rank tensor models are powerful in identifying abnormal traffic activity, they have multiple shortcomings. First, they rely on existing factorization models such as Tucker \cite{zhang2016tensor,xu2019anomaly} or CP \cite{li2020tensor}, which do not take the particular structure of anomalies, \textit{i.e.}, sparsity, into account when computing the low-dimensional projection.  Second, while higher-order robust principal component analysis \cite{geng2014high} takes the sparsity of anomalies into account, it  does not impose any additional structure such as  spatial and temporal contiguity on the detected anomalies.  Finally, existing approaches to quantify the spatiotemporal contiguity rely on total variation (TV)-norm of the sparse tensor component across spatial and temporal dimensions \cite{cao2015total,luo2023low}. While this formulation is well-suited for Euclidean data, it is not directly applicable to non-Euclidean data, i.e., tensors defined on a product graph. 


In this paper, we  propose a tensor-based anomaly detection approach that simultaneously takes into account the characteristics of normal and anomalous data. Extending our preliminary work in \cite{indibi2024spatiotemporal}, we model the normal activity as low-rank and anomalies as sparse, temporally persistent, \textit{i.e.}, the local changes last for a reasonably long time period, and  spatially smooth, \textit{i.e.}, neighboring regions are likely to exhibit similar anomalous activity.   This is accomplished by extending the traditional higher order robust  PCA (HoRPCA) framework \cite{goldfarb2014robust} with additional regularizers that impose geometric structure on the sparse part. The temporal persistence and spatial contiguity of the sparse part are quantified by minimizing the graph signal variation with respect to the underlying temporal and spatial graphs, respectively. This formulation introduces a generalization of  the conventional total variation norm to non-Euclidean domains by considering the appropriate graph gradient operators, i.e., Laplacians, across both spatial and temporal domains. After the sparse component, $\mathcal{S}$, is extracted, we propose an anomaly scoring method that  models the entries of $\mathcal{S}$ as a normal random variable whose mean and variance depend on its local neighborhood. Based on this assumption, we derive a likelihood function that quantifies the likelihood of an anomaly for a given location and time point. The proposed framework is evaluated on both simulated and real data.
\section{Background}\label{background}   
\subsection{Notations}
Calligraphic letters $\mathcal{X}$ denote tensors, uppercase bold letters $\mathbf{X}$ denote matrices, lowercase bold letters $\mathbf{x}$ denote vectors and plain lowercase or uppercase letters $(n\,\, \text{or}\,\,N)$ denote scalars. For an $N$-mode tensor $\mathcal{X}$, $x_{i_1,i_2,\ldots,i_N}$ is its $(i_1,i_2,\ldots,i_N)$-th element. For a matrix $\mathbf{X}$, $X_{ij}$ is used to denote its $(i,j)$-th element and for a vector $\mathbf{x}$, $x_i$ is its $i$-th element. For a set $S$, $|S|$ denotes the cardinality of the set. $[N]$ is used to represent the set $\{1,2,\ldots,N\}$. 
For a matrix $\mathbf{X}$, $\mathbf{X}_{i\cdot}$ denotes the $i$-th row and  $\mathbf{X}_{\cdot i}$ denotes the $i$-th column. Let $\mathcal{I}=\{j_1,j_2,\ldots,j_m\}\subseteq [N]$ be the subset of row indices of $\mathbf{X}$. $(\mathbf{X})_{\mathcal{I}, j:k}$ indicates a submatrix of $\mathbf{X}$ formed by selecting the rows corresponding to the index set $\mathcal{I}$ and the columns from $j$ to $k$. $\mathbf{I}$ and $\mathbf{0}$ are used to denote the identity matrix and the matrix of all zeros, respectively. $\mathbf{1}_{m\times n}$ denotes the matrix of all ones with $m$ rows and $n$ columns. The operators $\operatorname{max}(.,.)$, $\operatorname{min}(.,.)$, and $\operatorname{sgn}(.)$ are used for maximum, minimum, and sign functions, respectively. 
\vspace{-2em}
\subsection{Definitions}
\begin{definition}
(Mode-$n$ fibers) Mode-$n$ fibers of a tensor $\mathcal{X} \in \mathbb{R}^{I_1\times I_2 \times \ldots \times I_N}$, denoted by $\mathbf{x}_{i_1,\ldots,i_{n-1},:,i_{n+1},\ldots,i_N} \in \mathbb{R}^{I_n}$, are vectors obtained by fixing all indices except the $n$-th dimension.
\end{definition}

\begin{definition}
(Mode-$n$ unfolding) Mode-$n$ unfolding of a tensor $\mathcal{X} \in \mathbb{R}^{I_1\times I_2 \times \ldots \times I_N}$ is a matrix $\mathbf{X}_{(n)}\in \mathbb{R}^{I_n \times \prod_{i=1,i\neq n}^N I_i}$  obtained by arranging the mode-$n$ fibers of $\mathcal{X}$ into its columns.
\end{definition}
\begin{definition}
(Mode-$n$ product)  The mode-$n$ product of a tensor $\mathcal{X}\in \mathbb{R}^{I_1\times I_2\times\ldots I_N}$ with a matrix $\mathbf{U}\in \mathbb{R}^{J\times I_n}$ is the tensor $\mathcal{A} = \mathcal{X}\times_n \mathbf{U}$, where $\mathcal{A}\in \mathbb{R}^{I_1\times I_2\times \ldots I_{n-1}\times J \times I_{n+1}\times \ldots \times I_N}$, and each element is defined as $a_{i_1,\ldots,i_{n-1},j,i_{n+1},\ldots,i_N} = \sum_{i_n = 1}^{I_n} x_{i_1,\ldots,i_N}U_{ji_n}$. This product can also be represented in matrix form as $\mathbf{A}_{(n)} = \mathbf{UX}_{(n)}$.
\end{definition}
\begin{definition}
(Tensor rank) Although there are multiple notions of tensor rank, in this paper, we use the Tucker rank. For a tensor $\mathcal{X}\in \mathbb{R}^{I_1\times I_2\times \ldots I_N}$, the Tucker rank is given by $\text{rank}_T(\mathcal{X}) = (\text{rank}_1(\mathcal{X}), \text{rank}_2(\mathcal{X}),\ldots,\text{rank}_N(\mathcal{X}) )$, where $\text{rank}_n(\mathcal{X}) = \text{rank}(\mathbf{X}_{(n)})$.
\end{definition}
\begin{definition}
(Tensor Norms) For an $N$-mode tensor $\mathcal{X}\in \mathbb{R}^{I_1\times I_2\times \ldots \times I_N}$, the $\ell_p$-norm is defined as $\|\mathcal{X}\|_p = \left(\sum_{i_1,i_2,\ldots,i_N} |x_{i_1,i_2,\ldots,i_N}|^p\right)^{1/p}$. The Frobenius norm is a special case with $p=2$, given by $\|\mathcal{X}\|_F$. The nuclear norm of the tensor $\mathcal{X}$ can be defined  as the weighted sum of the nuclear norms (SNN) \cite{liu2012tensor} of the mode-$n$ unfoldings $\mathbf{X}_{(n)}$ as $\|\mathcal{X}\|_* = \sum_{n=1}^N \psi_n \|\mathbf{X}_{(n)}\|_*$, where $\psi_n > 0$ is the weight assigned to the $n$-th mode and $\|\mathbf{X}_{(n)}\|_* = \sum_{j=1}^{\min (I_n,\prod_{i=1,i\neq n}^N I_i) } \sigma_j(\mathbf{X}_{(n)})$, where $\sigma_j(\mathbf{X}_{(n)})$ is the $j$-th singular value of $\mathbf{X}_{(n)}$.
\end{definition}

\begin{definition}
(Frobenius inner product) Given $\mathbf{A}\in \mathbb{R}^{m\times n}$ and $\mathbf{B} \in \mathbb{R}^{m\times n}$, the Frobenius inner product of $\mathbf{A}$ and $\mathbf{B}$ is $\langle\mathbf{A}, \mathbf{B}\rangle_F= \sum_{i\in[m]}\sum_{j\in[n]}A_{ij}B_{ij}$.
\end{definition}

\begin{definition}
(Hadamard product) Given $\mathbf{A}\in \mathbb{R}^{m\times n}$ and $\mathbf{B} \in \mathbb{R}^{m\times n}$, the Hadamard product of $\mathbf{A}$ and $\mathbf{B}$ is $\mathbf{C} = \mathbf{A}\circ \mathbf{B}\in \mathbb{R}^{m\times n}$ , where $C_{ij} = A_{ij}B_{ij}, \forall i\in [m], j \in [n]$.
\end{definition}

\begin{definition}
(Kronecker product) The Kronecker product of $\mathbf{A}\in \mathbb{R}^{m\times n}$ and $\mathbf{B}\in \mathbb{R}^{p\times q}$, denoted by $\mathbf{A}\otimes \mathbf{B} \in \mathbb{R}^{mp \times nq}$, is given by, 
\vspace{-1em}

\[
\setlength{\arraycolsep}{2pt}      
\renewcommand{\arraystretch}{0.8}   
\mathbf{A} \otimes \mathbf{B} =
\begin{bmatrix}
A_{11}\mathbf{B} & A_{12}\mathbf{B} & \ldots & A_{1n}\mathbf{B} \\
A_{21}\mathbf{B} & A_{22}\mathbf{B} & \ldots & A_{2n}\mathbf{B} \\
\vdots & \vdots & \vdots & \vdots \\
A_{m1}\mathbf{B} & A_{m2}\mathbf{B} & \ldots & A_{mn}\mathbf{B}
\end{bmatrix}.
\]

\end{definition}

\begin{definition}
(Undirected graph) An undirected binary graph is represented  as $G=(V, E, \mathbf{A})$ where $V$ is the node set with cardinality $|V|=n$, $E$ is the edge set \cite{newman2018networks}. The adjacency matrix, $\mathbf{A} \in \setR^{n\times n}$, with $A_{ij} = A_{ji}$ is defined as 
\vspace{-0.5em}$$A_{ij} = \begin{cases}
1, & \text{if } (v_i, v_j) \in E, \\
0, & \text{otherwise}.
\end{cases}$$ 
\end{definition}
\begin{definition}
(Graph Laplacian)  The graph Laplacian for graph $G$ is defined as $\mathbf{L} = \mathbf{D} - \mathbf{A}$, where $\mathbf{D}$ is the diagonal degree matrix with elements $D_{ii} = \sum_{j=1}^n A_{ij}$. The normalized Laplacian is defined as $\mathbf{L}_n = \mathbf{I} - \mathbf{D}^{-1/2}\mathbf{A}\mathbf{D}^{-1/2}$.

\end{definition}
\begin{definition} (Graph Total Variation)
A graph signal $\mathbf{x}\in \mathbb{R}^{n}$ 
is a vector whose entries
reside on the nodes of an unsigned graph, $G$. Signal smoothness quantifies
how much the signal samples vary with respect to the underlying
graph. In this paper,  
we use the graph total variation based on the $\ell_{p}$ norm \cite{chen2015signal}: $S_{p}(\mathbf{x})=\|\mathbf{x}
-\mathbf{A}_{n}\mathbf{x}\|_{p}^{p}$, 
where $\mathbf{A}_{n}$ is the normalized adjacency matrix to ensure that the
shifted signal is properly scaled with respect to the original one.
    
\end{definition}
\begin{definition}
(Discrete Differentiation Operator) The discrete differentiation operator is defined as a matrix $\mathbf{\Delta}\in\mathbb{R}^{n-1\times n}$, where, 
\vspace{-1em}

\[
\begingroup
\setlength{\arraycolsep}{2pt}      
\renewcommand{\arraystretch}{0.8}   
\mathbf{\Delta} =
\begin{bmatrix}
1 & -1 & 0 & \ldots & 0 & 0 \\
0 & 1 & -1 & \ldots & 0 & 0 \\
\vdots & \vdots & \vdots & \vdots & \vdots & \vdots \\
0 & 0 & 0 & \ldots & 1 & -1
\end{bmatrix}.
\endgroup
\]

\end{definition}

\begin{definition}($k$-NN graph)
Given a tensor $\mathcal{Y}\in \mathbb{R}^{I_1\times I_2\times \ldots\times I_N}$, let $G_n = (V^n, E^n, \mathbf{W}^n)$ be the graph corresponding to mode-$n$, where $V_n$ is the set of vertices (i.e., rows of $\mathbf{Y}_{(n)}$), $E^n$ is the set of edges, and $\mathbf{W}^n$ is the weighted adjacency matrix corresponding to mode-$n$.  $\mathbf{W}^n$ can be constructed as the $k$-NN graph with the Gaussian kernel, where $W^n_{s,s'}$ is defined by:

\vspace{-1em}

\begin{align*}
\resizebox{0.95\linewidth}{!}{$
W^n_{s,s'} =
\begin{cases}
\exp\!\left(-\dfrac{\left\| \mathbf{Y}_{(n)_{s\cdot}} - \mathbf{Y}_{(n)_{s'\cdot}} \right\|_F^2}{2\sigma^2}\right) & 
\text{if } \mathbf{Y}_{(n)_{s\cdot}} \in \mathcal{N}_k(\mathbf{Y}_{(n)_{s'\cdot}}) \\
& \text{or } \mathbf{Y}_{(n)_{s'\cdot}} \in \mathcal{N}_k(\mathbf{Y}_{(n)_{s\cdot}}), \\[1em]
0 & \text{otherwise},
\end{cases}
$}
\end{align*}

\noindent where $\mathcal{N}_k(\mathbf{Y}_{(n)_{s\cdot}})$ is the Euclidean $k$-nearest neighborhood of the $s$-th row of $\mathbf{Y}_{(n)}$ and $\sigma^2$ is the variance parameter of the Gaussian kernel. 
\end{definition}

\begin{definition}(Proximal operator)
The proximal operator of a proper convex function $f:\mathbb{R}^n\to \mathbb{R}\cup \{+\infty\}$, denoted $\mathbf{prox}_{f}:\mathbb{R}^n\to\mathbb{R}^n$ is defined by the relation

\[\mathbf{prox}_f(v)=\underset{x\in \mathbf{dom}f}{\mathrm{argmin}}\big(f(x)+\frac{1}{2}\|x-v\|^2\big)\]

\end{definition}

\section{Anomaly Feature Extraction}\label{method}
In the proposed method, we model the spatiotemporal data as a multi-way tensor $\calY\in\mathbb{R}^{I_1\times \ldots \times I_N}$ where different modes may correspond to spatial and temporal domains as well as different types of features. We make the following three assumptions about the characteristics of the anomalies to model our data. 
\begin{assumption}
    Normal activity can be modeled as a low-rank tensor while the anomalies are modeled as the sparse part, i.e., $\cal{Y}=\calX+\calS$.
\end{assumption}
This assumption is motivated by prior work in spectral methods for anomaly detection \cite{li2015low,wang2022learning}, where the normal activity is assumed to live in a low-dimensional subspace, whereas the anomalous activity cannot be represented by the span of that subspace.
\begin{assumption}
    The anomalies are assumed to last for periods of time, i.e., have strong short-term dependencies. 
\end{assumption}
 This assumption ensures that instantaneous changes in the data, which may be due to errors in sensing, are not mistaken for actual anomalies. This assumption can be quantified by the graph total variation in the time domain with $p=1$ as follows:
 
 \begin{align}
    S_1(\mathbf{S}_{(t)})&=  \|\mathbf{S}_{(t)} - \mathbf{A}_{t} \mathbf{S}_{(t)}\|_1^1
    =\|\mathbf{\Delta} \mathbf{S}_{(t)}\|_1 = \|\calS \times_t \mathbf{\Delta} \|_1,
    \label{eq:timetotalvariation}
\end{align}

where the first order discrete time differentiation operator, $\Delta$, is  equivalent to the Laplacian of the time domain line graph, i.e.,  $\Delta=\mI-\mA_{t}$, with $\mA_{t}$ being the cyclic time shift operator defined in \cite{ortega2018graph}.

 \begin{assumption}
Group anomalies within spatiotemporal data exhibit themselves as spatially contiguous groups of locations, i.e., the local variation of the anomalies is sparse. 
 \end{assumption}
 This assumption states that if there is an anomaly at a particular location, its neighbors should also exhibit anomalous activity. This assumption can be quantified by minimizing the variation of the anomaly  with respect to the spatial graph, $\mA_{s}$, with $p=1$.
In our case, the graph signals with respect to the spatial domain are the column vectors of the unfolding of $\calS$ across the location mode, $l$.

\begin{align}
    S_1(\mathbf{S}_{(l)})&=  \|\mathbf{S}_{(l)} - \mathbf{A}_n \mathbf{S}_{(l)}\|_1=
    \|\mL_n \mathbf{S}_{(l)}\|_1 = \| \calS \times_l \mL_n \|_1.
\end{align}

Considering these three assumptions, we formulate the spatiotemporal anomaly detection problem as follows: 

\begin{mini}|l|[0]
{\calX,\calS} 
{\sum_{i=1}^N (\psi_{i}\|\mX_{(i)}\|_*)+\lambda_1\|\mathcal{S}\|_1} 
{}{}
\breakObjective{+\lambda_l\|\mathcal{S}\times_l \mathbf{L}_n\|_1 +\lambda_t\|\mathcal{S}\times_t \Delta\|_1\label{LR_STSS}}
\addConstraint{\calX+\calS}{=\calY}{},
\end{mini}

\noindent where $\lambda_{1},\lambda_{l}$ and $\lambda_{t}$ are the regularization parameters that control the level of sparsity, amount of spatial contiguity and amount of temporal contiguity, respectively. 

\subsection{Optimization Algorithm}
The optimization problem in \eqref{LR_STSS} involves minimizing sum of multiple convex penalties for the variables $\calX, \calS$ jointly, and does not have a closed form solution. We iteratively solve this problem using a two block Alternating Direction Method of Multipliers (ADMM) \cite{boyd2011distributed}. ADMM allows us to decouple the dependencies between the different variables and alternatingly minimize for the variables using proximal operators in an iterative fashion.

For terms containing $\mathcal{S}$, we introduce auxiliary variables $\mathcal{W=S}$, $\mathcal{W}_l = \mathcal{S}\times_l \mathbf{L}_n$, $\mathcal{W}_t=\mathcal{S}\times_t \Delta$. Similarly, we introduce auxiliary variables $\calX_{1}=\calX_2=\ldots=\calX_{N} = \calX$ to separate the dependencies between the sum of nuclear norms containing the mode unfoldings of $\calX$. The resulting optimization problem can be written as,

\begin{align}
\underset{\calX, \{\calX_i\}_{i=1}^N, \calS, \calW, \calW_l, \calW_t
}{\text{minimize}} &\,\,\,\, \sum_{i=1}^N \psi_{i}\|\mathbf{X}_{i_{(i)}}\|_* +\lambda_1\|\calW\|_1 \label{eq:LR-STSS_admm}\\
 &+\lambda_l\|\calW_l\|_1  +\lambda_t\|\calW_t\|_1 \notag \\
\text{subject to}\,\,\, & \calX=\calX_i, \quad \text{for } i=1,...,N \notag\\
&\calX+\calS=\calY,  \calW=\calS,\notag\\
 &\calW_l= \calS\times_l\mathbf{L}_n,  \calW_t=\calS \times_t\Delta. \notag
\end{align}


The augmented Lagrangian of the problem with step size $\rho$ is expressed as \eqref{eq:LR-STSS_lagrangian} where $\{\Lambda_i\}_{i=1}^N$,$\Lambda_f$,$\Lambda$, $\Lambda_t$, $\Lambda_l$ are the dual variables for the equality constraints $\{\calX=\calX_i\}_{i=1}^N$, $\calX+\calS=\calY$, $\calW=\calS$, $\calW_t=\calS\times_t\mathbf{\Delta}$, $\calW_l \times_l \mathbf{L}_n$ respectively.

\begin{align}
    \calL_\rho=\sum_{i=1}^N \big(&\psi_{i}\|\mathbf{X}_{i_{(i)}}\|_* + \frac{\rho}{2}\|\calX -\calX_i + \frac{\Lambda_i}{\rho}\|_F^2\big) \label{eq:LR-STSS_lagrangian} \\
    &+ \lambda_1 \|\calW\|_1+  \frac{\rho}{2}\|\calW - \calS + \frac{\Lambda}{\rho}\|_F^2 \nonumber\\
    &+ \lambda_l \|\calW_l\|_1+ \frac{\rho}{2}\|\calW_l - \calS\times_l\mathbf{L}_n + \frac{\Lambda_l}{\rho}\|_F^2 \nonumber\\
    &+ \lambda_t \|\calW_t\|_1+ \frac{\rho}{2}\|\calW_t - \calS\times_t\mathbf{\Delta} + \frac{\Lambda_t}{\rho}\|_F^2, \notag \\
    & + \frac{\rho}{2} \|\calX+\calS-\calY +\frac{\Lambda_f}{\rho}\|_F^2. \notag
\end{align}

\noindent Following the two block ADMM scheme, the variables are grouped into two blocks, namely $\{\calX, \calW, \calW_t, \calW_l\}$ and $\{\calX_1,\ldots,\calX_N, \calS\}$. The augmented Lagrangian $\eqref{eq:LR-STSS_lagrangian}$ is alternatingly minimized for the variables in each block, followed by the dual variable updates as in Algorithm \ref{alg:lr_stss}.

Minimizing \eqref{eq:LR-STSS_lagrangian}, the update steps for the variables $\calW, \calW_l, \calW_t$ reduce to the proximal mappings of $\ell_1$ norm defined as $\mathbf{prox}_{\lambda\|.\|_1}(\mathcal{B}) = \mathrm{sign}(\mathcal{B})\circ\mathrm{max}(| \mathcal{B} |-\lambda,0)$. The update steps for the variables $\calX_1,\ldots,\calX_N$ reduce to the proximal operator of the Schatten-1 norm, obtained by the singular value thresholding operation $\mathbf{prox}_{\psi\|\cdot\|_*}(\mB)=\mathcal{SVT}(\mB,\psi)$. The update for $\calX$ reduces to a simple averaging operation. Finally, the update step for the variable $\calS$ involves the quadratic minimization problem.
\vspace{-1.5em}

\begin{align}
    &\mathrm{min}_\calS \; \frac{\rho}{2}\|\calX^{k+1}+\calS-\calY + \frac{\Lambda_f^k}{\rho}\|_F^2 \nonumber \\
    &+ \frac{\rho}{2}\|\calW_t^{k+1}-\calS\times_t \mathbf{\Delta}+\frac{\Lambda_t^{k}}{\rho}\|_F^2 \nonumber\\
    &+ \frac{\rho}{2}\|\calW_l^{k+1} - \calS\times \mL_n + \frac{\Lambda_l^k}{\rho}\|_F^2 + \frac{\rho}{2}\|\calW^{k+1} - \calS + \frac{\Lambda^k}{\rho}\|_F^2.\label{eqn:s_update_problem}
    \vspace{-2.5em}
\end{align}

The minimizer of the problem in \eqref{eqn:s_update_problem} is equivalent to the solution of the linear system of equations, $\mG \mS^{k+1}_{(l,t)}=\mB^k_{(l,t)}$, where $\calS^{k+1}$ is matricized by mapping the location $(l)$ and time $(t)$ modes onto the columns of $\mS_{(l,t)}^{k+1}$ and $\mB^k_{(l,t)}$ is the matricization of the tensor $\calB^k$. The tensor $\calB^k$ and the matrix $\mG$ are given as follows:

\begin{align*}
 \calB^k=& \Big(\big(\Bt\big)\times_t \mathbf{\Delta}^T  + \big(\Bl \big)\times_l \mL_n^T \\
 & +\big(\Bone\big) +  \big( \Bb\big) \Big),\\
 \mathbf{G}=& [\mI_l \otimes (\Delta^T\Delta+ \mI_t) + (\mL_n^T \mL_n + \mI_l)\otimes \mI_t].
\end{align*}

Since the Laplacian $\mL_n$ and $\Delta$ are sparse matrices, this step can be solved efficiently using conjugate gradient descent. In our implementation, we calculate the solution by caching the eigendecomposition of matrices $(\mL_n^T\mL_n + \mI_l)=\Phi_l \mD_l \Phi_l^T$ and $( \mathbf{\Delta}^T\mathbf{\Delta} + \mI_t)=\Phi_t \mD_l \Phi_l^T$. The inverse matrix admits the eigen-decomposition $\mG^{-1}= (\Phi_l \otimes \Phi_t) \mD_\oplus^{-1} (\Phi_l\otimes\Phi_t)^T$ where $\mD_\oplus= \mD_l\otimes \mI_t + \mI_l\otimes\mD_t$. We use this fact to implement the inverse product  $\mS^{k+1}_{(l,t)} =\mG^{-1}{\mB^k}_{(l,t)}$ using mode products and elementwise products working on the tensor $\calB$. 
\section*{LR-STSS Pseudocode}

\begin{algorithm}
\caption{Low-rank and spatiotemporally smooth anomaly separation (LR-STSS)}
\label{alg:lr_stss}
\begin{algorithmic}[1]  
\Procedure{LR-STSS}{$\calY, \mL_n, \lambda_1, \lambda_t, \lambda_l, \psi_i, \rho$}
    \State Initialize $\calX, \calS, \calW, \calW_t, \calW_l \gets 0$
    \For{$i = 1,\ldots,N$}
        \State $\calX_i \gets 0$  \Comment{Primal variables}
        \State $\Lambda_i \gets 0$  \Comment{Dual variables}
    \EndFor
    \State Initialize $\Lambda, \Lambda_f, \Lambda_t, \Lambda_l \gets 0$

    \For{$k = 0, \ldots, \mathrm{maximum\_iteration}$}
        
        \State $\calX^{k+1} \gets (\sum_i (\calX_i^k + \Lambda_i^k / \rho) + (\calY - \calS^k - \Lambda_f^k / \rho)) / (N+1)$
        \State $\calW_t^{k+1} \gets \text{prox}_{\lambda_t / \rho} (\calS^k \times_t \mathbf{\Delta} - \Lambda_t^k / \rho)$
        \State $\calW_l^{k+1} \gets \text{prox}_{\lambda_l / \rho} (\calS^k \times_l \mL_n - \Lambda_l^k / \rho)$
        \State $\calW^{k+1} \gets \text{prox}_{\lambda_1 / \rho} (\calS^k - \Lambda_l^k / \rho)$

        \For{$i = 1,\ldots,N$}
            \State $\mX_i^{k+1} \gets \text{prox}_{\psi_i / \rho} (\mX_i^{k} + \Lambda_i^k / \rho)$
        \EndFor
        \State $\mS^{k+1} \gets \mG^{-1} \mB^k$

        \State $\Lambda^{k+1} \gets \Lambda_l^k + \rho (\calW^{k+1} - \calS^{k+1})$
        \State $\Lambda_f^{k+1} \gets \Lambda_f^k + \rho (\calX^{k+1} + \calS^{k+1} - \calY)$
        \State $\Lambda_t^{k+1} \gets \Lambda_t^k + \rho (\calW_t^{k+1} - \calS^{k+1} \times_t \mathbf{\Delta})$
        \State $\Lambda_l^{k+1} \gets \Lambda_l^k + \rho (\calW_l^{k+1} - \calS^{k+1} \times_l \mL_n)$
        \For{$i = 1,\ldots,N$}
            \State $\Lambda_i^{k+1} \gets \Lambda_i^k + \rho (\calX^{k+1} - \calX_i^{k+1})$
        \EndFor
    \EndFor
\EndProcedure
\end{algorithmic}
\end{algorithm}

\subsection{Convergence}\label{global_conv}
The convergence of Algorithm \ref{alg:lr_stss}  can be established by  showing that the proposed optimization problem \eqref{eq:LR-STSS_admm} can be formulated as a two-block ADMM problem and using prior research in the global convergence of non-convex optimization problems \cite{deng2016global}. 
\begin{theorem}
The sequence $\{\calX^k, \calS^k\}$ generated by Algorithm \ref{alg:lr_stss} converges globally  to the optimal solution of Eq. \eqref{eq:LR-STSS_admm}. \label{thm:global-conv}
\end{theorem}
\vspace{-0.5em}
\noindent
The proof of the theorem is given in Appendix \ref{app:global-conv}.

\section{Anomaly Scoring}\label{anomaly_scoring}
In this section, following \cite{zheng2024graph} we propose a method for anomaly scoring by calculating the anomaly likelihood of each entry of the estimated sparse tensor $\hat{\mathcal{S}} \in\mathbb{R}^{I_{1}\times I_{2}\times \ldots\times I_{N}}$. The anomaly likelihood provides information about how anomalous the current location and time point is with respect to its neighborhood. 

Without loss of generality, we assume that modes-$1$ and $2$ correspond to the location and time modes, respectively. Let  $T=\prod_{i=1}^N I_i, T_2=\prod_{i=2}^N I_i$ and $T_{3}=\prod_{i=3}^N I_i$. For each element $s\in [I_1]$ and $j\in [T_2]$, we model $\hat{S}_{(1)_{s,j}}$ as

\begin{align}
\hat{S}_{(1)_{s,j}} &\sim \mathcal{N}(\mu_{s,j}, \sigma^2_{s,j}),\,\,\, s\in[I_1],\,\,j\in \left[T_2\right]\label{scoring:model}
\end{align}

\noindent where  $\mu_{s,j}$ and $\sigma^2_{s,j}$ are given by,

\begin{align}
\mu_{s,j} &= 
 \frac{\sum_{b\in [I_2]}
\left\langle \mathbf{B}^{(s,b)}, \mathbf{w}^{(s,b)}\mathbf{1}_{1\times T_3}
\right\rangle_F}
{T_3\sum_{b\in[I_2]}\|\mathbf{w}^{(s,b)}\|_1},
  \label{scoring:mu}\\
\sigma_{s,j}^2 &=
\left[ \frac{\sum_{b\in [I_2]}
\left\langle \mathbf{B}^{(s,b)}\circ \mathbf{B}^{(s,b)}, 
\mathbf{w}^{(s,b)}\mathbf{1}_{1\times T_3}\right\rangle_F}
{T_3\sum_{b\in[I_2]}\|\mathbf{w}^{(s,b)}\|_1}
- \mu^2_{s,j} \right], \label{scoring:sigma}
\end{align}


\noindent 
where $\mathbf{B}^{(s,b)}\!\!\in\!\! \mathbb{R}^{n_s^k\times T_3}$ is a submatrix of  $\hat{\mathbf{S}}_{(1)}$, defined for each time point and the  neighborhood of location $s$ as $\mathbf{B}^{(s,b)}\!\!=\!\! (\hat{\mathbf{S}}_{(1)})_{N_k(s):1+(b-1)T_3 : bT_3}$, with $s \in [I_{1}],b\in [I_2]$. Without loss of generality, let us assume that the first row of $\mathbf{B}^{(s,b)}$ corresponds to the center node $s$ and \(N_{k}(s)=\{u\in V\mid d(s,u)\leq k\}\) with $n^{k}_{s}=|N_{k}(s)|$ is the local $k$-hop neighborhood of node $s$ defined using the underlying spatial graph, $G_{s}$, and shortest path distance $d(s,u)$. $\mathbf{w}^{(s,b)}\in \mathbb{R}^{n^k_s\times 1}$ is the vector of weights, 
where for $i\in[n^k_s]$, $w^{(s,b)}_i$ is given by,

\begin{align}
w^{(s,b)}_i = \exp\left(-\frac{\left\| \bar{\mathbf{B}}^{(s,b)}_{i\cdot} - \bar{\mathbf{B}}^{(s,b)}_{1\cdot} \right\|_F^2}{2\tau^2}\right),\label{scoring:weight}
\end{align}

\noindent 
where $\tau$ is a tuning parameter and $\bar{\mathbf{B}}^{(s,b)}$ is defined as,

\begin{align}
\vspace{-2em}
\bar{\mathbf{B}}^{(s,b)} = \begin{cases} \mathbf{B}^{(s,b)},& \text{if}\,\, b = 1,\\[0.5em]
\begin{bmatrix} \mathbf{B}^{(s,b-1)} & \mathbf{B}^{(s,b)}\end{bmatrix}, & \text{ if} \,\, b = I_2, \\[0.5em]
\begin{bmatrix} \mathbf{B}^{(s,b-1)} & \mathbf{B}^{(s,b)}& \mathbf{B}^{(s,b+1)}\end{bmatrix}, & \text{ otherwise.}
\label{scoring:st_block}
\end{cases}
\end{align}

$\bar{\mathbf{B}}^{(s,b)}$ is formed by augmenting $\mathbf{B}^{(s,b)}$  with its immediate temporal neighbors to incorporate temporal continuity. The weights in  \eqref{scoring:weight} quantify the spatiotemporal dependency between the center node and its spatial neighbors within a time window that includes immediate temporal neighbors of the current time block. Higher weights are assigned to the neighbors with higher similarity to the center node by effectively combining spatial contiguity and temporal persistence. 

We obtain anomaly likelihood by computing the negative log-likelihood of $\hat{S}_{(1)_{s,j}}$ for $s\in [I_1], j\in [T_2]$,

\begin{align}
l(\hat{S}_{(1)_{s,j}}) = \log(\sigma_{s,j}) + \frac{1}{2}\log(2\pi) + \frac{1}{2}\bigg(\frac{\hat{S}_{(1)_{s,j}} - \mu_{s,j}}{\sigma_{s,j}}\bigg)^2.\label{likelihood}
\end{align}

The negative log-likelihood  determines how unlikely the occurrence of the current value is, given its spatiotemporal neighborhood. Finally, for a chosen level of significance $\alpha\in (0,1)$ we determine the threshold $\gamma$ as the upper-$\alpha$ quantile of the negative log-likelihood distribution. All the points with anomaly scores exceeding this threshold $\gamma$ are selected as anomalies with $(1-\alpha)\times 100\%$ level of confidence.

\section{Results}\label{results}
In this section, we present results on both synthetic and real data to evaluate the effectiveness of the proposed method  in comparison to HoRPCA \cite{li2015low}, which induces sparsity but does not impose spatial and temporal smoothness on the sparse components. 
For synthetic data experiments, the performance of all methods is quantified using average AUC-ROC and F1 scores over 50 iterations calculated based on the anomaly scoring method introduced in Section \ref{anomaly_scoring}, referred to as the NLL scoring method. In real data experiments, we compare NLL scoring with the absolute scores $|\hat{\mathcal{S}}|$, referred to as the ABS scoring method. We denote the AUC-ROC computed using NLL scoring as NLL-AUC-ROC, and that computed using the ABS scoring method as ABS-AUC-ROC.

\subsection {Hyperparameter Selection}
The performance of the methods depends on the choice of hyperparameters $\{\psi_i\}, \lambda_1, \lambda_l, \lambda_t $, which control the different aspects of the data. Python library Optuna \cite{akiba2019optuna} is used to select the hyperparameters over a pre-determined range. We select $\lambda_1$ in $(0,1)$ and set $\psi_i = 1-\lambda_1, \forall i$. This sampling scheme for $\lambda_1$ and $\psi_i$ enables us to systematically explore all possible values of the ratio $\lambda_1/\psi_i$, which controls the balance between the level of sparsity and low-rankness of the data. $\lambda_l$ and $\lambda_t$ are sampled from the range between $10^{-8}$ and $10$ using a log-uniform distribution to ensure exploration of both very large and small values. For experiments where the ground truth is available, these hyperparameters are selected by maximizing the AUC-ROC and F1 scores simultaneously. For real data without ground truth, e.g., 2018 NYC Taxi data, we select the hyperparameters to maximize the number of top-$3\%$ detected events.
\subsection{Synthetic Data Experiment}
For simulated data, an anomalous tensor $\mathcal{Y}\in \mathbb{R}^{40\times 24 \times 7 \times 20}$ is constructed as $\mathcal{Y} = \mathcal{X}+\mathcal{S}+\mathcal{E}$, where $\mathcal{X}$ represents the low-rank tensor corresponding to normal activity, $\mathcal{S}$ is the sparse tensor corresponding to anomalies, and $\mathcal{E}$ is  additive white Gaussian noise. To generate $\cal X$, we specify the Tucker rank as $\text{rank}_T(\mathcal{X})=(8,8,5,5)$. Next, we create a random core tensor $\mathcal{C}\in \mathbb{R}^{8\times 8 \times 5 \times 5}$ whose entries are drawn from the standard normal distribution. Finally, we generate random orthonormal matrices $\mathbf{U}_i\in\mathbb{R}^{m_i \times n_i}, i\in [4]$, and construct $\mathcal{X}$ as $\mathcal{X}= \mathcal{C}\times_1 \mathbf{U}_1 \times_2 \mathbf{U}_2 \times_3 \mathbf{U}_3 \times_4 \mathbf{U}_4$ and normalize it. The sparse tensor $\mathcal{S}$ is constructed based on the underlying spatiotemporal graph $G$, which is set to be a 2-dimensional $8\times 5$ grid graph where the nodes are connected to their immediate horizontal and vertical neighbors. Random anomaly centers for each anomaly group are generated by randomly selecting tensor coordinates, indexed by $(i_1, i_2, i_3, i_4)$, where $i_1$ denotes the spatial center and $i_4$ denotes the temporal center. All the vertices within an $r$-hop neighborhood of each local anomaly center $i_1$ are set to be an anomaly, where $r$ denotes the spatial radius of the anomaly. A rectangular pulse with duration $d$ centered around index $i_4$ is applied to each of these locations to generate the spatiotemporal anomaly with temporal persistence. 
\subsubsection{\textbf{Ablation Study}}
\begin{figure*}[t]
    \centering
    \includegraphics[width=0.85\textwidth]{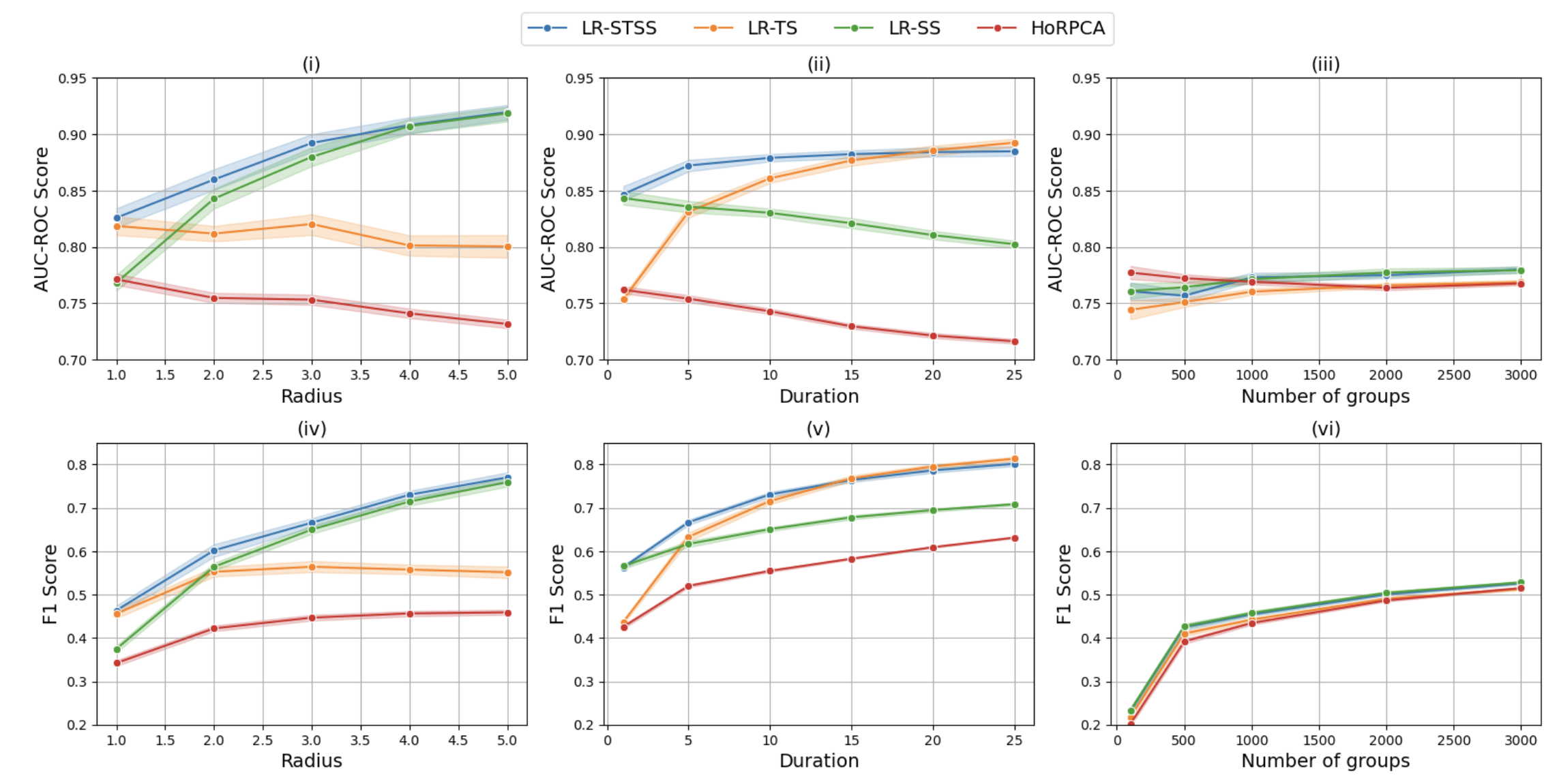}
    \caption{Ablation study showing the impact of different components of the model. AUC-ROC scores are reported for varying (i) Radius (with $d = 4$ and $g = 100$), (ii) Duration (with $r = 2$, $g = 450$), and (iii) Number of groups of anomalies (with $r = 1$ and $d = 1$). F1 scores are reported for varying (iv) Radius, (v) Duration, and (vi) Number of groups of anomalies with same parameter settings as in AUC-ROC experiments. SNR and amplitude are fixed at 10 and 0.25, respectively. Shaded regions correspond to the $95\%$ confidence intervals calculated over 50 iterations.} 
    \vspace{-1em}
    \label{fig:ablation_study}
\end{figure*}
In this section, we perform an ablation study by systematically removing the different regularization terms inducing spatial and temporal smoothness from the proposed model \eqref{LR_STSS}, Low-Rank Spatio-temporally Smooth (LR-STSS), to understand their effect under different scenarios. We denote the model when the spatial regularization is removed ($\lambda_l = 0$)  as Low-Rank Temporally Smooth (LR-TS), when the temporal component is omitted ($\lambda_t = 0$) as Low-Rank Spatially Smooth (LR-SS) and finally, we have HoRPCA when $\lambda_l = \lambda_t  = 0$. In Fig. \ref{fig:ablation_study}, we report AUC-ROC and F1 scores for all four aforementioned methods as the anomaly radius $(r)$, duration $(d)$ and number of anomaly groups $(g)$ are varied with amplitude of anomaly $(c)$ set to $0.25$ and signal-to-noise ratio (SNR) fixed at $10$ dB. When the spatial radius of the anomalies is increased (Fig. \ref{fig:ablation_study} (i), (iv)), LR-STSS and LR-SS perform better. This is expected as these methods incorporate the assumption of spatial contiguity  into the extraction of $\mathcal{S}$.  The absence of this term leads to poor performance for LR-TS and HoRPCA. LR-STSS performs best when $r\leq4$, after which the performances of LR-STSS and LR-SS become comparable, with LR-SS achieving a slightly higher AUC-ROC score. When the temporal duration of the anomalies increases (Fig. \ref{fig:ablation_study} (ii), (v)), we observe higher AUC-ROC and F1 scores for LR-STSS and LR-TS compared to LR-SS and HoRPCA, highlighting the importance of the temporal contiguity in the model. For $d\leq 15$, LR-STSS performs best, after which point  LR-TS performs better as it emphasizes temporal smoothness. HoRPCA performs the worst in both scenarios since it lacks both the spatial and temporal smoothness components. When the radius and duration are held constant, changing the number of groups of anomalies (Fig. \ref{fig:ablation_study} (iii), (vi)) does not change the average AUC-ROC scores of different methods. Although all the methods are more or less comparable, LR-STSS and LR-SS perform slightly better than LR-TS and HoRPCA. In Fig. \ref{fig:ablation_study} ((i), (ii), (iv), (v)), we also observe that even in the cases where AUC-ROC score has a decreasing trend, F1 score continues to  increase. This phenomenon can be explained by the fact that the AUC-ROC score promotes the quality of global ranking as it summarizes the performance of the model over all possible thresholds. On the other hand, F1 score reflects a local quality of classification at a particular threshold. Therefore, if the False Positive rate increases, AUC-ROC might decrease, but the F1 score may still increase if the chosen threshold effectively balances the precision and recall.
\subsubsection{\textbf{Robustness Against Noise}}
\begin{figure*}[t]
    \centering
    \includegraphics[width=0.8\textwidth]{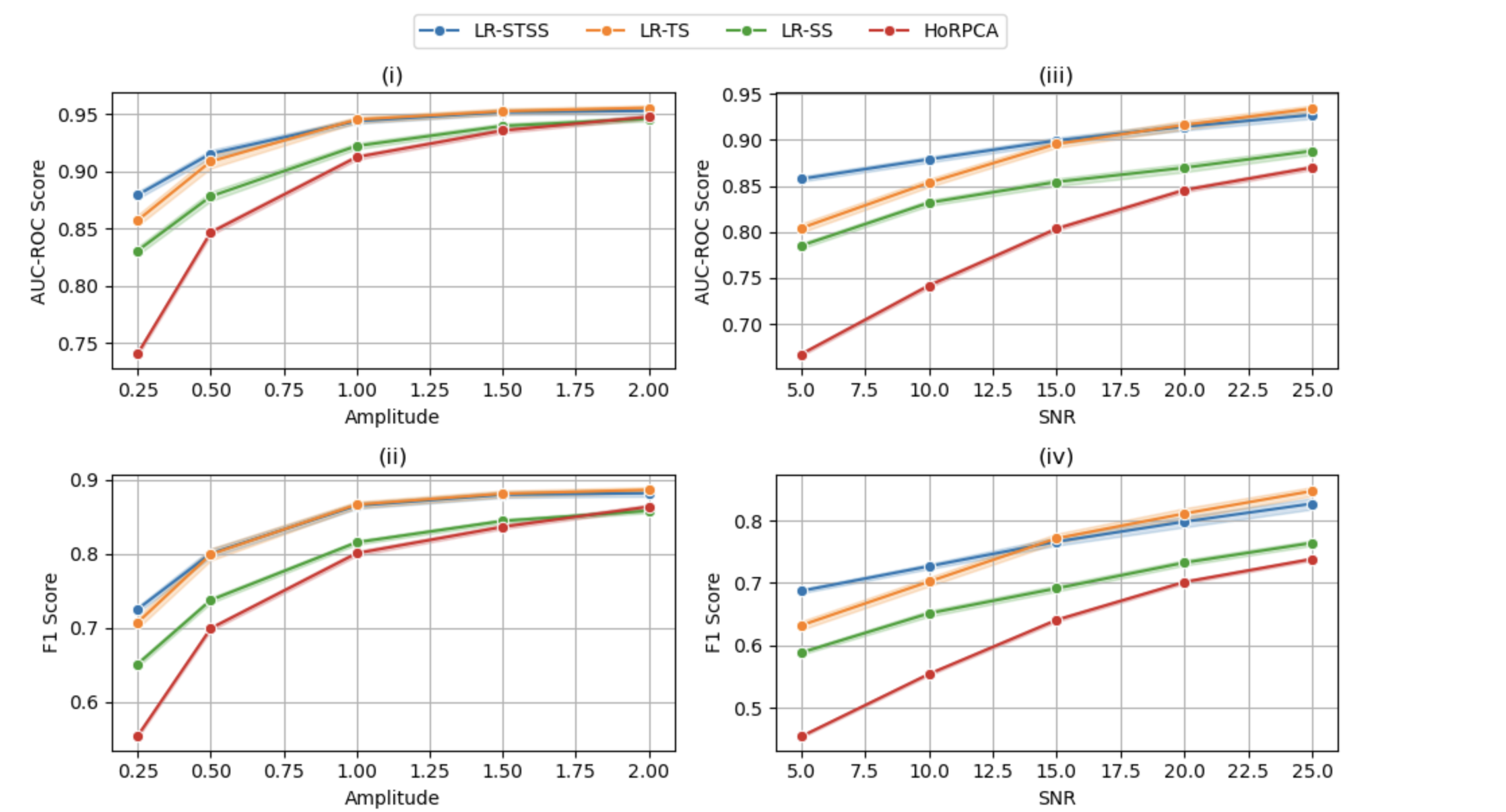}
    \caption{Analysis of robustness against various methods. AUC-ROC scores for (i) varying amplitude (SNR = $10$), (ii) varying SNR ($c = 0.25$) and F1 scores for (iii)  varying amplitude (with SNR = $10$) and (iv) varying SNR (with $c = 0.25$) are reported. Radius, duration and number of groups are fixed at 2, 10 and 450 respectively. Shaded regions correspond to 95$\%$ confidence intervals over 50 iterations.}
    \vspace{-1em}
    \label{fig:noise_study}
\end{figure*}
In this section, we vary the amplitude of the anomaly and SNR to assess the robustness of the methods with fixed radius ($r=2$), duration ($d=10$) and number of anomaly groups ($g=450$). Amplitude  determines the strength of the anomaly; lower amplitude makes it harder to distinguish anomalies from noise.  AUC-ROC and F1 scores in  Fig. \ref{fig:noise_study} (i)-(ii) indicate that for varying amplitude, LR-STSS and LR-TS achieve the highest level of performance. Due to the effect of high temporal persistence ($d$ = 10), LR-TS slightly outperforms LR-STSS when the anomaly strength increases above 1. Among all the methods, HoRPCA performs the worst, especially when the anomaly amplitude is small. Fig. \ref{fig:noise_study} (iii)-(iv)  depict a similar trend where for lower SNR, LR-STSS performs the best. As SNR increases, LR-TS outperforms LR-STSS due to long temporal duration. However, LR-STSS maintains a comparable performance. On the other hand, HoRPCA exhibits the worst performance as compared to the other three methods.

\subsection{2018 NYC Taxi Data}
The first real dataset is NYC yellow taxi trip records for 2018 \footnote{\url{https://www1.nyc.gov/site/tlc/about/tlc-trip-record-data.page}}. This dataset consists of trip information such as the departure zone and time, arrival zone and time, number of passengers, and trips for each yellow taxi trip in NYC. In the following experiments, we only use the arrival zone and time to collect the number of arrivals for each zone aggregated over one-hour time intervals. We selected 81 central zones to avoid zones with very low traffic. Thus, we created a tensor, $\mathcal{Y}\in \mathbb{R}^{24\times 7 \times 52 \times 81}$ where the first mode corresponds to hours within a day, the second mode corresponds to days of a week, the third mode corresponds to weeks of a year and the last mode corresponds to the zones. To evaluate the performance of the proposed methods on real data, we compiled a list of 20 urban events \cite{sofuoglu2022gloss} that took place in the important urban activity centers such as city squares, parks, museums, stadiums and concert halls, during 2018. 
 
Table \ref{table:nyc} reports the number of detected events among 20 compiled events  for varying top-$K\%$ of anomaly scores using NLL  and ABS scoring methods, respectively. We observe that LR-STSS can detect more events for smaller top-$\alpha\%$, followed by LR-TS. LR-SS and HoRPCA exhibit poor detection performance, which emphasizes the importance of incorporating temporal information along with spatial information in the model. Comparing the two anomaly scoring methods in Table \ref{table:nyc}, we observe that for event detection using LR-STSS and HoRPCA, the proposed NLL scoring method performs better than the ABS scoring method. Both scoring methods yield comparable results for LR-TS and LR-SS.  


\begin{table}[H]
\centering
\renewcommand{\arraystretch}{0.8} 
\setlength{\tabcolsep}{0.8pt} 
\small 
\resizebox{0.49\textwidth}{!}{%
\begin{tabular}{l*{8}{|cc}}
\toprule
\textbf{Top}-$\mathbf{K\%}$
 & \multicolumn{2}{c}{0.014} & \multicolumn{2}{c}{0.07} & \multicolumn{2}{c}{0.14} & \multicolumn{2}{c}{0.3} 
 & \multicolumn{2}{c}{0.7}  & \multicolumn{2}{c}{1}    & \multicolumn{2}{c}{2}    & \multicolumn{2}{c}{3} \\
\cmidrule(lr){2-3}\cmidrule(lr){4-5}\cmidrule(lr){6-7}\cmidrule(lr){8-9}
\cmidrule(lr){10-11}\cmidrule(lr){12-13}\cmidrule(lr){14-15}\cmidrule(lr){16-17}
\textbf{Method} 
 & \text{NLL} & \text{ABS} & \text{NLL} & \text{ABS} & \text{NLL} & \text{ABS} & \text{NLL} & \text{ABS}
 & \text{NLL} & \text{ABS} & \text{NLL} & \text{ABS} & \text{NLL} & \text{ABS} & \text{NLL} & \text{ABS} \\
\midrule
\midrule
LR-STSS & 3 & 3 & 5 & 5 & 9 & 7 & 12 & 11 & 15 & 15 & 16 & 16 & 19 & 18 & 19 & 19 \\
LR-TS   & 2 & 2 & 4 & 4 & 4 & 4 & 6 & 6 & 8 & 8 & 11 & 11 & 13 & 13 & 17 & 17 \\
LR-SS   & 1 & 1 & 1 & 1 & 1 & 1 & 2 & 2 & 3 & 3 & 4 & 3 & 7 & 7 & 13 & 13 \\
HoRPCA  & 0 & 0 & 0 & 0 & 1 & 2 & 5 & 2 & 7 & 2 & 8 & 3 & 11 & 7 & 15 & 10 \\
\bottomrule
\end{tabular}

}
\caption{Number of detected events among 20 compiled events in NYC dataset for varying top-$K$\% scores under NLL and ABS scoring.}
\label{table:nyc}

\end{table}

\subsection{Server Machine Dataset}
Server Machine Dataset (SMD) \cite{su2019robust} is a collection of multivariate time-series metrics collected from an internet company, with anomaly labels provided by domain experts based on incident reports. SMD is comprised of $28$ different time-series datasets, which we refer to as the channels, corresponding to 3 distinct machines with 8, 9, 11 channels, respectively. Dataset in each channel is split into training and testing parts where the anomaly labels are only available for the testing part. Each channel has a varying number of time-points available, corresponding to $38$ variates every consecutive minute. In total, there are 708420 time points, of which $4.16\%$  are anomalous. We apply our methods only using the test datasets. We process the channels for each machine  by creating 4-mode tensors, $\mathcal{Y}_m\in \mathbb{R}^{38\times {d_m}\times 24\times 60}$, for $m\in [28]$, where the first mode corresponds to the number of features, the second mode to the number of days for the $m$-th machine channel, the third mode to the number of hours in a day, and the fourth mode to the number of minutes in an hour. Since the datasets vary by duration, the last day is incomplete. We completed these missing entries with the mean across other days. We consider the mode of features as the spatial domain and construct a binary $2$-NN graph using the Euclidean distance between the 38 features.

Table \ref{table:smd} presents the mean and standard deviation of NLL-AUC-ROC and ABS-AUC-ROC scores across $28$ machines for different anomaly detection methods, evaluated using two different scoring algorithms. We observe that LR-TS followed by LR-STSS achieves the highest AUC-ROC among all four methods, indicating that for the server machine dataset, models with temporal contiguity play a more important role in capturing anomalies. HoRPCA exhibits the worst performance as it incorporates neither the spatial nor the temporal contiguity. NLL-AUC-ROC outperforms ABS-AUC-ROC across all four methods, highlighting the effectiveness of the NLL scoring approach.
\begin{table}[H]

    \centering
    \renewcommand{\arraystretch}{0.8}
    \resizebox{0.3\textwidth}{!}{%
    \begin{tabular}{c|c|c}
    \midrule
         \textbf{Method} & \textbf{NLL-AUC-ROC } & \textbf{ABS-AUC-ROC} \\
         \midrule
         \midrule
         LR-STSS & $0.841 \pm 0.113$ & $0.800 \pm 0.124$ 
         \\
         LR-TS & $0.870 \pm 0.115$ & $0.825 \pm 0.148$ 
         \\
         LR-SS & $0.838 \pm 0.107$ & $0.793 \pm 0.139$
         \\
         HoRPCA & $0.834 \pm 0.105$ & $ 0.766 \pm 0.167$
         \\
         \midrule
    \end{tabular}
    }
    \caption{AUC-ROC comparison across different methods and different scoring algorithms applied to SMD.}
    \label{table:smd}
    
\end{table}

\section{Conclusion}
In this paper, we proposed a regularized robust tensor decomposition for spatiotemporal anomaly detection, where we assume that normal activity is modeled as low-rank while the anomalies are modeled as sparse, spatially contiguous, and temporally persistent.  These assumptions are captured by graph total variation of the sparse part of the tensor with respect to the underlying temporal and spatial graphs. We further introduced a spatiotemporal anomaly scoring method that determines how unlikely the occurrence of an event is given its neighborhood.  The performance of the proposed algorithm is evaluated on both synthetic data and real data, where LR-STSS, LR-STS, and LR-SS consistently outperform HoRPCA. Although the proposed scoring method did not yield significant improvements in the synthetic data settings over absolute scoring, it exhibited noticeable gains in the real data experiments. All the implementation details are provided in: \url{https://github.com/Rachita-Mondal/Anomaly_Detection}. Future work will consider extensions of the proposed framework by incorporating higher-order dependencies in time and space using different graph operators. Moreover, the proposed models can be easily adapted to impute missing values by simultaneously completing the  low-rank part and detecting the anomalies.



\appendices
\section{Proof of Theorem \ref{thm:global-conv}}\label{app:global-conv}
We begin proving Theorem \ref{thm:global-conv} by first establishing the following lemma. 

Let us define the functions $f,g: \mathbb{R}^{\prod_{i=1}^N I_i} \rightarrow \mathbb{R}$ as,
\vspace{-0.7em}
\begin{align}
f(\mathbf{u}) &= \sum_{i=1}^N \big(\psi_{i}\|\mathbf{X}_{{i(i)}}\|_*\big),\notag \\
g(\mathbf{v}) &= \lambda_1\|\calW\|_1 + \lambda_l\|{\calW}_l\|_1 +\lambda_t\|{\calW}_t\|_1, \label{conv: obj}
\end{align}
where 

\[
\mathbf{u} = \begin{bmatrix}\mathbf{x}^T \!\!&\!\! \mathbf{w}^T \!\!&\!\! \mathbf{w}^T_{t} \!\!&\!\! \mathbf{w}^T_{l}\end{bmatrix}^T \,\,\text{and}\,\,
\mathbf{v} = \begin{bmatrix}\mathbf{s}^T \!\!&\!\! \mathbf{x}^T_{1} \!\!& \ldots \!\!& \mathbf{x}^T_{N}\end{bmatrix}^T
\] with $\mathbf{x} \!\!=\!\! \operatorname{vec}({\mathbf{X}_{(l)}})$, $\mathbf{w} \!\!=\!\! \operatorname{vec}({\mathbf{W}_{(l)}}), \mathbf{w}_{t}\!\! =\!\!\operatorname{vec}({\mathbf{W}_{t_{(l)}}}), \mathbf{w}_{l} \!\!=\!\! \operatorname{vec}({\mathbf{W}_{l_{(l)}}}), 
 \mathbf{s}\! = \!\operatorname{vec}({\mathbf{S}_{(l)}})$ and $\mathbf{x}_{i} \!= \!\operatorname{vec}({\mathbf{X}_{i_{(l)}}}), i\in [N]$. 
\begin{lemma}\label{lemma:two-block-ADMM}
The optimization problem in \eqref{eq:LR-STSS_admm} can be written in the form of a two-block ADMM problem as:
\vspace{-0.5em}
\begin{align}
\operatorname*{minimize}_{\mathbf{u},\mathbf{v}} \quad f(\mathbf{u}) + g(\mathbf{v}), \,\,
\textrm{subject to} \quad \mathbf{Au} + \mathbf{Bv} = \mathbf{c}. \label{ADMM-form}
\end{align}

\end{lemma}
\vspace{-1.95em}
\begin{proof}
Let us consider,
\vspace{-1em}
\begingroup
\setlength{\arraycolsep}{1.5pt}      
\renewcommand{\arraystretch}{0.8} 
{\footnotesize
\begin{align}
\mathbf{A} &= \begin{bmatrix}
    \mathbf{I} & \mathbf{I} & \ldots &\mathbf{I} & \mathbf{0} & \mathbf{0} & \mathbf{0}\\
    \mathbf{0} & \mathbf{0} & \ldots & \mathbf{0} & \mathbf{I} & \mathbf{0} & \mathbf{0}\\
    \mathbf{0} & \mathbf{0} & \ldots & \mathbf{0} & \mathbf{0} & \mathbf{0} & \mathbf{I}\\
    \mathbf{0} & \mathbf{0} & \ldots & \mathbf{0} & \mathbf{0} & \mathbf{P} & \mathbf{0}
\end{bmatrix}^T,
\mathbf{B}  = \begin{bmatrix}
\mathbf{I} & \mathbf{0} & \mathbf{0} & \ldots &\mathbf{0} & -\mathbf{I} &  \mathbf{D} & \mathbf{E}\\
\mathbf{0} &-\mathbf{I} & \mathbf{0} & \ldots & \mathbf{0} & \mathbf{0} & \mathbf{0} & \mathbf{0}\\
\mathbf{0} &\mathbf{0} & -\mathbf{I} & \ldots & \mathbf{0} & \mathbf{0} & \mathbf{0} & \mathbf{0}\\
\vdots & \vdots & \vdots & \vdots & \vdots & \vdots & \vdots\\
\mathbf{0} &\mathbf{0} & \mathbf{0} & \ldots & -\mathbf{I} & \mathbf{0} & \mathbf{0} & \mathbf{0}
    \end{bmatrix}^T \label{conv: constraints}
    \end{align}
    }
\endgroup
\noindent
\noindent
where $\mathbf{D} = -(\mathbf{I} \otimes\mathbf{\Delta})\mathbf{Q}$ and $\mathbf{E} = -(\mathbf{I} \otimes \mathbf{L}_n) $. $\mathbf{P}$ and $\mathbf{Q}$ are the appropriate permutation matrices such that $\mathbf{w}_l = \mathbf{P} \mathbf{w}_t$ and $\mathbf{s} = \mathbf{Q} \operatorname{vec}(\mathbf{S}_{(t)})$. Using the definitions of $f$ and $g$ in (\ref{conv: obj}), along with ${\mathbf{A}}$ and $\mathbf{B}$ from \eqref{conv: constraints}, the optimization problem \eqref{eq:LR-STSS_admm} can be rewritten as a two-block ADMM problem in the form of \eqref{ADMM-form} with $\mathbf{c}=[\operatorname{vec}(\calY)^T\;\;\mathbf{0}^T]^T$.
\end{proof}
Since the functions $f$ and $g$ are closed, proper, and convex, using Theorem 2.2 of \cite{deng2016global}, it can be shown that any sequence  $\{\calX^k, \calS^k\}$ generated by the ADMM algorithm globally converges to the optimal solution.


\begin{thebibliography}{10}
\providecommand{\url}[1]{#1}
\csname url@samestyle\endcsname
\providecommand{\newblock}{\relax}
\providecommand{\bibinfo}[2]{#2}
\providecommand{\BIBentrySTDinterwordspacing}{\spaceskip=0pt\relax}
\providecommand{\BIBentryALTinterwordstretchfactor}{4}
\providecommand{\BIBentryALTinterwordspacing}{\spaceskip=\fontdimen2\font plus
\BIBentryALTinterwordstretchfactor\fontdimen3\font minus \fontdimen4\font\relax}
\providecommand{\BIBforeignlanguage}[2]{{%
\expandafter\ifx\csname l@#1\endcsname\relax
\typeout{** WARNING: IEEEtran.bst: No hyphenation pattern has been}%
\typeout{** loaded for the language `#1'. Using the pattern for}%
\typeout{** the default language instead.}%
\else
\language=\csname l@#1\endcsname
\fi
#2}}
\providecommand{\BIBdecl}{\relax}
\BIBdecl

\bibitem{bgfg_sep_smooth}
B.~Shen, R.~R. Kamath, H.~Choo, and Z.~Kong, ``Robust tensor decomposition based background/foreground separation in noisy videos and its applications in additive manufacturing,'' \emph{IEEE Transactions on Automation Science and Engineering}, vol.~20, no.~1, pp. 583--596, 2023.

\bibitem{yan2014image}
H.~Yan, K.~Paynabar, and J.~Shi, ``Image-based process monitoring using low-rank tensor decomposition,'' \emph{IEEE Transactions on Automation Science and Engineering}, vol.~12, no.~1, pp. 216--227, 2014.

\bibitem{zhang2016tensor}
X.~Zhang, G.~Wen, and W.~Dai, ``A tensor decomposition-based anomaly detection algorithm for hyperspectral image,'' \emph{IEEE Transactions on Geoscience and Remote Sensing}, vol.~54, no.~10, pp. 5801--5820, 2016.

\bibitem{tan2016short}
H.~Tan, Y.~Wu, B.~Shen, P.~J. Jin, and B.~Ran, ``Short-term traffic prediction based on dynamic tensor completion,'' \emph{IEEE Transactions on Intelligent Transportation Systems}, vol.~17, no.~8, pp. 2123--2133, 2016.

\bibitem{ju2012fuzzy}
Z.~Ju and H.~Liu, ``Fuzzy gaussian mixture models,'' \emph{Pattern Recognition}, vol.~45, no.~3, pp. 1146--1158, 2012.

\bibitem{rousseeuw2003robust}
P.~J. Rousseeuw and A.~M. Leroy, \emph{Robust regression and outlier detection}.\hskip 1em plus 0.5em minus 0.4em\relax John wiley \& sons, 2003.

\bibitem{gornitz2013toward}
N.~G{\"o}rnitz, M.~Kloft, K.~Rieck, and U.~Brefeld, ``Toward supervised anomaly detection,'' \emph{Journal of Artificial Intelligence Research}, vol.~46, pp. 235--262, 2013.

\bibitem{ruff2019deep}
L.~Ruff, R.~A. Vandermeulen, N.~G{\"o}rnitz, A.~Binder, E.~M{\"u}ller, K.-R. M{\"u}ller, and M.~Kloft, ``Deep semi-supervised anomaly detection,'' \emph{arXiv preprint arXiv:1906.02694}, 2019.

\bibitem{villa2021semi}
M.~E. Villa-P{\'e}rez, M.~A. Alvarez-Carmona, O.~Loyola-Gonz{\'a}lez, M.~A. Medina-P{\'e}rez, J.~C. Velazco-Rossell, and K.-K.~R. Choo, ``Semi-supervised anomaly detection algorithms: A comparative summary and future research directions,'' \emph{Knowledge-Based Systems}, vol. 218, p. 106878, 2021.

\bibitem{lin2018anomaly}
C.~Lin, Q.~Zhu, S.~Guo, Z.~Jin, Y.-R. Lin, and N.~Cao, ``Anomaly detection in spatiotemporal data via regularized non-negative tensor analysis,'' \emph{Data Mining and Knowledge Discovery}, vol.~32, no.~4, pp. 1056--1073, 2018.

\bibitem{zhang2018detecting}
H.~Zhang, Y.~Zheng, and Y.~Yu, ``Detecting urban anomalies using multiple spatio-temporal data sources,'' \emph{Proceedings of the ACM on interactive, mobile, wearable and ubiquitous technologies}, vol.~2, no.~1, pp. 1--18, 2018.

\bibitem{chae2012spatiotemporal}
J.~Chae, D.~Thom, H.~Bosch, Y.~Jang, R.~Maciejewski, D.~S. Ebert, and T.~Ertl, ``Spatiotemporal social media analytics for abnormal event detection and examination using seasonal-trend decomposition,'' in \emph{2012 IEEE conference on visual analytics science and technology (VAST)}.\hskip 1em plus 0.5em minus 0.4em\relax IEEE, 2012, pp. 143--152.

\bibitem{wu2008spatio}
E.~Wu, W.~Liu, and S.~Chawla, ``Spatio-temporal outlier detection in precipitation data,'' in \emph{International workshop on knowledge discovery from sensor data}.\hskip 1em plus 0.5em minus 0.4em\relax Springer, 2008, pp. 115--133.

\bibitem{li2020tensor}
Z.~Li, N.~D. Sergin, H.~Yan, C.~Zhang, and F.~Tsung, ``Tensor completion for weakly-dependent data on graph for metro passenger flow prediction,'' in \emph{proceedings of the AAAI conference on artificial intelligence}, vol.~34, no.~04, 2020, pp. 4804--4810.

\bibitem{xu2019anomaly}
M.~Xu, J.~Wu, H.~Wang, and M.~Cao, ``Anomaly detection in road networks using sliding-window tensor factorization,'' \emph{IEEE Transactions on Intelligent Transportation Systems}, vol.~20, no.~12, pp. 4704--4713, 2019.

\bibitem{kasai2016network}
H.~Kasai, W.~Kellerer, and M.~Kleinsteuber, ``Network volume anomaly detection and identification in large-scale networks based on online time-structured traffic tensor tracking,'' \emph{IEEE Transactions on Network and Service Management}, vol.~13, no.~3, pp. 636--650, 2016.

\bibitem{papalexakis2018network}
E.~E. Papalexakis, A.~Beutel, and P.~Steenkiste, ``Network anomaly detection using co-clustering,'' in \emph{Encyclopedia of Social Network Analysis and Mining}.\hskip 1em plus 0.5em minus 0.4em\relax Springer, 2018, pp. 1501--1516.

\bibitem{geng2014high}
X.~Geng, K.~Sun, L.~Ji, and Y.~Zhao, ``A high-order statistical tensor based algorithm for anomaly detection in hyperspectral imagery,'' \emph{Scientific reports}, vol.~4, p. 6869, 2014.

\bibitem{cao2015total}
X.~Cao, L.~Yang, and X.~Guo, ``Total variation regularized {RPCA} for irregularly moving object detection under dynamic background,'' \emph{IEEE transactions on cybernetics}, vol.~46, no.~4, pp. 1014--1027, 2015.

\bibitem{luo2023low}
Y.~Luo, X.~Zhao, Z.~Li, M.~K. Ng, and D.~Meng, ``Low-rank tensor function representation for multi-dimensional data recovery,'' \emph{IEEE transactions on pattern analysis and machine intelligence}, vol.~46, no.~5, pp. 3351--3369, 2023.

\bibitem{indibi2024spatiotemporal}
M.~Indibi and S.~Aviyente, ``Spatiotemporal group anomaly detection via graph total variation on tensors,'' in \emph{ICASSP 2024-2024 IEEE International Conference on Acoustics, Speech and Signal Processing (ICASSP)}.\hskip 1em plus 0.5em minus 0.4em\relax IEEE, 2024, pp. 7035--7039.

\bibitem{goldfarb2014robust}
D.~Goldfarb and Z.~Qin, ``Robust low-rank tensor recovery: Models and algorithms,'' \emph{SIAM Journal on Matrix Analysis and Applications}, vol.~35, no.~1, pp. 225--253, 2014.

\bibitem{liu2012tensor}
J.~Liu, P.~Musialski, P.~Wonka, and J.~Ye, ``Tensor completion for estimating missing values in visual data,'' \emph{IEEE transactions on pattern analysis and machine intelligence}, vol.~35, no.~1, pp. 208--220, 2012.

\bibitem{newman2018networks}
M.~Newman, \emph{Networks}.\hskip 1em plus 0.5em minus 0.4em\relax Oxford university press, 2018.

\bibitem{chen2015signal}
S.~Chen, A.~Sandryhaila, J.~M. Moura, and J.~Kova{\v{c}}evi{\'c}, ``Signal recovery on graphs: Variation minimization,'' \emph{IEEE Transactions on Signal Processing}, vol.~63, no.~17, pp. 4609--4624, 2015.

\bibitem{li2015low}
S.~Li, W.~Wang, H.~Qi, B.~Ayhan, C.~Kwan, and S.~Vance, ``Low-rank tensor decomposition based anomaly detection for hyperspectral imagery,'' in \emph{2015 IEEE International Conference on Image Processing (ICIP)}.\hskip 1em plus 0.5em minus 0.4em\relax IEEE, 2015, pp. 4525--4529.

\bibitem{wang2022learning}
M.~Wang, Q.~Wang, D.~Hong, S.~K. Roy, and J.~Chanussot, ``Learning tensor low-rank representation for hyperspectral anomaly detection,'' \emph{IEEE Transactions on Cybernetics}, vol.~53, no.~1, pp. 679--691, 2022.

\bibitem{ortega2018graph}
A.~Ortega, P.~Frossard, J.~Kova{\v{c}}evi{\'c}, J.~M. Moura, and P.~Vandergheynst, ``Graph signal processing: Overview, challenges, and applications,'' \emph{Proceedings of the IEEE}, vol. 106, no.~5, pp. 808--828, 2018.

\bibitem{boyd2011distributed}
S.~Boyd, N.~Parikh, E.~Chu, B.~Peleato, J.~Eckstein \emph{et~al.}, ``Distributed optimization and statistical learning via the alternating direction method of multipliers,'' \emph{Foundations and Trends{\textregistered} in Machine learning}, vol.~3, no.~1, pp. 1--122, 2011.

\bibitem{deng2016global}
W.~Deng and W.~Yin, ``On the global and linear convergence of the generalized alternating direction method of multipliers,'' \emph{Journal of Scientific Computing}, vol.~66, pp. 889--916, 2016.

\bibitem{zheng2024graph}
Y.~Zheng, H.~Y. Koh, M.~Jin, L.~Chi, H.~Wang, K.~T. Phan, Y.-P.~P. Chen, S.~Pan, and W.~Xiang, ``Graph spatiotemporal process for multivariate time series anomaly detection with missing values,'' \emph{Information Fusion}, vol. 106, p. 102255, 2024.

\bibitem{akiba2019optuna}
T.~Akiba, S.~Sano, T.~Yanase, T.~Ohta, and M.~Koyama, ``Optuna: A next-generation hyperparameter optimization framework,'' in \emph{Proceedings of the 25th ACM SIGKDD international conference on knowledge discovery \& data mining}, 2019, pp. 2623--2631.

\bibitem{sofuoglu2022gloss}
S.~E. Sofuoglu and S.~Aviyente, ``Gloss: Tensor-based anomaly detection in spatiotemporal urban traffic data,'' \emph{Signal Processing}, vol. 192, p. 108370, 2022.

\bibitem{su2019robust}
Y.~Su, Y.~Zhao, C.~Niu, R.~Liu, W.~Sun, and D.~Pei, ``Robust anomaly detection for multivariate time series through stochastic recurrent neural network,'' in \emph{Proceedings of the 25th ACM SIGKDD international conference on knowledge discovery \& data mining}, 2019, pp. 2828--2837.

\end{thebibliography}
\bibliographystyle{IEEEtran}   

\end{document}